\documentclass[runningheads]{llncs}
\usepackage{graphicx}
\usepackage{tikz}
\usepackage{comment}
\usepackage{amsmath,amssymb} % define this before the line numbering.
\usepackage{color}
\usepackage{wrapfig}
\usepackage{multirow}

\usepackage[accsupp]{axessibility}  % Improves PDF readability for those with disabilities.

\begin{document}
% \linenumbers
\pagestyle{headings}
\mainmatter

\title{Geometry-aware Single-image Full-body Human Relighting} % Replace with your title

%\begin{comment}
% \titlerunning{ECCV-22 submission ID \ECCVSubNumber} 
% \authorrunning{ECCV-22 submission ID \ECCVSubNumber} 
% \author{Anonymous ECCV submission}
% \institute{Paper ID \ECCVSubNumber}
%\end{comment}
%******************

% \begin{comment}
\author{Chaonan Ji\inst{1} \and Tao Yu\inst{1} \and
Kaiwen Guo\inst{2} \and Jingxin Liu\inst{3} \and Yebin Liu\inst{1}}
\authorrunning{C. Ji et al.}
\institute{Department of Automation, Tsinghua University, China \and
Meta Reality Labs \and Guangdong OPPO Mobile Telecommunications Corp., Ltd}
% \end{comment}
%******************
\maketitle

\begin{abstract}
Single-image human relighting aims to relight a target human under new lighting conditions by decomposing the input image into albedo, shape and lighting. Although plausible relighting results can be achieved, previous methods suffer from both the entanglement between albedo and lighting and the lack of hard shadows, which significantly decrease the realism. To tackle these two problems, we propose a geometry-aware single-image human relighting framework that leverages single-image geometry reconstruction for joint deployment of traditional graphics rendering and neural rendering techniques. For the de-lighting, we explore the shortcomings of UNet architecture and propose a modified HRNet, achieving better disentanglement between albedo and lighting. For the relighting, we introduce a ray tracing-based per-pixel lighting representation that explicitly models high-frequency shadows and propose a learning-based shading refinement module to restore realistic shadows (including hard cast shadows) from the ray-traced shading maps. Our framework is able to generate photo-realistic high-frequency shadows such as cast shadows under challenging lighting conditions. Extensive experiments demonstrate that our proposed method outperforms previous methods on both synthetic and real images.
\end{abstract}

% \vspace{-0.8cm}
\section{Introduction}
Human relighting aims to relight a target human with correct shadow effects under a desired illumination. Relighting can realize seamless replacement of the background while ensuring the light consistency of the foreground and the background and has a huge application prospect in film-making, video chat and Virtual Reality. 
Single-image human relighting for a general human person is convenient and promising for amateur photographers but also more challenging because the difficulty of decoupling lighting, geometry, and surface material from a single image. 

Most existing single-image human relighting methods focus on portrait relighting \cite{pandey2021total,sun2019single,wang2020single,nagano2019deep,barron2014shape,egger2018occlusion,genova2018unsupervised,lin2020towards,nestmeyer2020learning,sengupta2018sfsnet,shahlaei2015realistic,shu2017neural,tewari2017mofa,li2018closed,shu2017portrait,zhou2019deep,zhang2020portrait,hou2021towards} and only a few works \cite{Manuel,Kanamori,tajimaPG21} focus on single-image full-body relighting. 
%  Compared with single image portrait relighting, we have two key observations for difficulties in single image human relighting. 
Since the style, texture and material of human clothes are widely varied and the geometry and poses of clothed humans are usually complex, decoupling the albedo and lighting is highly ill-posed. Moreover, mutual occlusion of limbs will produce self-shadows, which are not only difficult to remove for de-lighting tasks, but also challenging to generate under the target lighting conditions for relighting tasks. Previous full-body relighting methods \cite{Manuel,Kanamori,tajimaPG21} have attempted to address these issues, but the results still have the following drawbacks: (i) inability to disambiguate albedo and lighting (ii) inability to model high-frequency shadows due to reliance on Spherical harmonics representation of lighting. 
% We tackle these two problems in this paper.

\begin{figure}[t]
\centering
\setlength{\abovecaptionskip}{0.cm}
\includegraphics[width=0.99\textwidth]{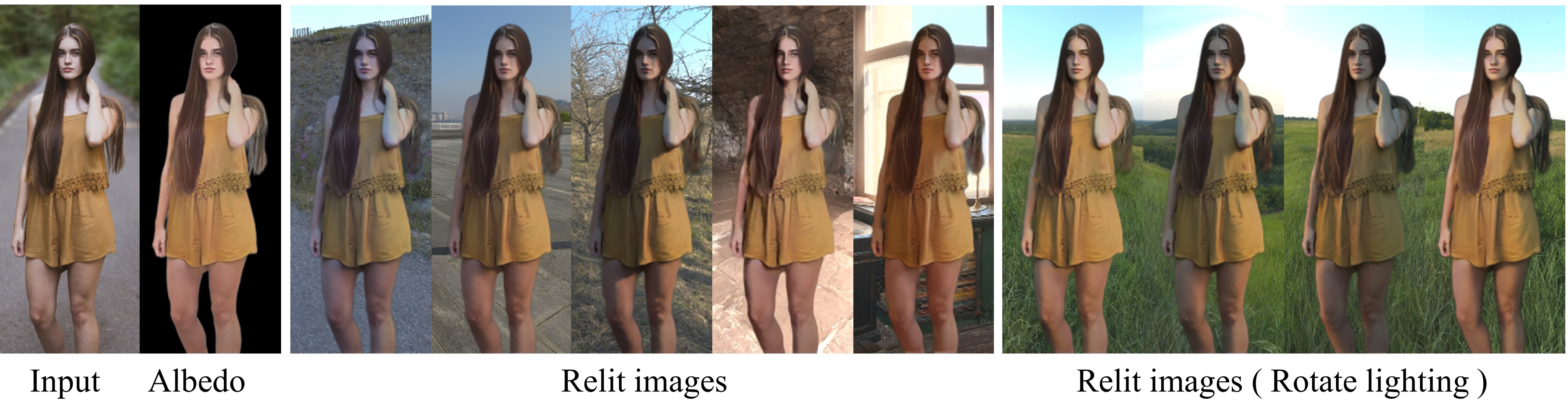}
\caption{Given a single human image and an arbitrary high dynamic range lighting environment, our framework estimates a high-quality albedo map and generates photorealistic relit images of the target human subject under the desired lighting conditions.}
\label{fig:ablation_refinement}
% \vspace{-0.6cm}
\end{figure}

We propose a novel geometry-aware single-image human relighting framework that leverages SOTA single-image geometry reconstruction \cite{saito2020pifuhd,guo2020towards} for fully leveraging the
%the joint deployment of
traditional graphics rendering and neural rendering techniques simultaneously.
%  especially for challenging lighting conditions. 
First of all, accurate intrinsic decomposition for albedo estimation (de-lighting) is the cornerstone of producing high-quality relighting results. Previous methods \cite{Manuel,Kanamori,tajimaPG21} have used the UNet to infer albedo and suffer from severe entanglement between lighting and albedo. 
We discover that the skip connections in UNet is the culprit and found that HRNet\cite{sun2019deep} performs much better for de-lighting. 
In addition we further improve the de-lighting performance by: i) removing the aggressive down-sampling operations in the early stage, ii) eliminating skip connections and iii) fusing multi-scale features while maintaining high-resolution representations in the HRNet, and finally achieve better disentanglement between lighting and albedo on both synthetic and real images. 
%  We demonstrate the validity of our delighting method in experiments.
 
%  but vanilla HRnet struggle to generate high-resolution albdeo prediction result. Vanilla HRnet with skip connection solves the problem of albedo blur, but introduces the lighting information from input to output, causing part of the input image's shadows to remain on the predicted albedo. 
 
More importantly, even with high-quality albedo, we still need photo-realistic shadows to produce realistic relit results. 
Limited by the expressive ability of the spherical harmonics (SH) lighting model and the lack of 3D geometry information, previous methods \cite{Manuel,Kanamori,tajimaPG21} can only produce low-frequency shadows. Low-resolution environment maps \cite{sun2019single,wang2020single} and pre-filtered environment maps \cite{pandey2021total} have also been explored and struggle to generate hard cast shadows. 
With the development of single-image 3D human reconstruction technology, obtaining high-quality mannequins from a single image is possible, which can provide more 3D prior information for human relighting. Hence, we propose a geometry-aware relighting method, which consists of a ray tracing-based per-pixel lighting representation that explicitly models high-frequency shadows and a learning-based shading refinement module to restore realistic shadows. 
The key idea of our method is that with the estimated 3D human model, we can render photo-realistic shading maps that encode full-band lighting information under target lighting conditions using ray tracing. This lighting representation is able to preserve high-frequency shadows and significantly improve the quality of the relit results which remains difficult for learning-based methods. However, the ray-traced shadows still suffer from artifacts due to the errors in geometry estimation, which prevents direct composition of the relit results. 
% To eliminate shadow errors caused by inaccurate geometry estimation in the ray tracing step, 
Thus, we further propose a learning-based refinement module that utilizes the ray-traced shading maps and the inferred ambient occlusion map as input to restore photo-realistic high-frequency shadows with rich local details and clear shadow boundaries. The proposed framework dramatically improves relighting performance and produces a photo-realistic relit image with high-frequency self-shadowing effects under the target lighting conditions. 
 
To conclude, our main contributions are the following:

\begin{itemize}
\setlength{\parsep}{0pt}
\setlength{\parskip}{0pt}
\item A novel geometry-aware single-image human relighting framework that combines single-image 3D human reconstruction, ray tracing and neural rendering technologies.
\item We demonstrate the UNet with skip connections is not suitable for the delighting task and propose a modified HRNet for achieving better disentanglement between lighting and albedo on both synthetic images and real photographs. 
\item We propose a ray tracing-based per-pixel lighting representation and a learning-based shading refinement module that utilizes the inferred ambient occlusion map as auxiliary input to restore photo-realistic shadows while preserving rich local details. 
\end{itemize}

% \vspace{-0.4cm}
\section{Related Work}

\subsection{Person-Specific Human Relighting}
Turner Whitted \cite{whitted1979improved} first used recursive ray tracing to simulate global illumination and render realistic images under target lighting conditions. However, the forward rendering technique requires a high-precision 3d model and corresponding PBR textures of the target object, which are unavailable or costly for real-world objects. Debevec \textit{et al.}\cite{debevec2000acquiring} proposed collecting OLAT(one-look-at-a-time) images using LightStage to synthesize specific person’s face under novel illuminations with no 3D model required. Subsequent works improved LightStage to capture higher-quality images and a larger range of human bodies\cite{hawkins2001photometric,debevec2002lighting,wenger2005performance,debevec2012light,weyrich2006analysis,chabert2006relighting,guo2019relightables,zhang2021neural}. Guo \textit{et al.}\cite{guo2019relightables} attempted to combine image-based rendering technique with geometry and material esrimated by LightStage and achieved unprecedented quality and photorealism for free viewpoint videos. However, collecting OLAT data, gradient data and object geometry requires professional equipment and it is a time-consuming and complicated process for person-specific relighting. Li \textit{et al.}\cite{li2013capturing} attempted to relight target human from multi-view video recorded under unknown illumination conditions and Imber \textit{et al.}\cite{imber2014intrinsic} extended this approach to scene relighting by introducing intrinsic textures. These two methods can produce high-quality relit results but multiview video must be collected for every target human or scene.

\subsection{General Human Relighting}
Deep learning makes general human relighting possible. Most existing general human relighting methods\cite{pandey2021total,sun2019single,wang2020single,nagano2019deep,barron2014shape,egger2018occlusion,genova2018unsupervised,lin2020towards,nestmeyer2020learning,sengupta2018sfsnet,shahlaei2015realistic,shu2017neural,tewari2017mofa,li2018closed,shu2017portrait,zhou2019deep,zhang2020portrait,hou2021towards,meka2019deep} focus on human portrait relighting. \cite{sun2019single,zhang2021neural,pandey2021total} achieved amazing single-image portrait relighting utilizing a dataset synthesized from OLAT images. For human full-body relighting, Meka \textit{et al.}\cite{meka2020deep} combined traditional geometry pipelines with neural rendering to generate relit results using gradient images and estimated human geometry. Kanamori \textit{et al.}\cite{Kanamori} proposed an occlusion-aware single-image full-body human relighting method to infer albedo, geometry and illumination leveraging spherical harmonics (SH) lighting model. On the basis of \cite{Kanamori}, Lagunas \textit{et al.}\cite{Manuel} added specular reflectance and light-dependent residual terms to explicitly handle highlights and Tajima \textit{et al.}\cite{tajimaPG21} used a residual network to restore neglected light effects. However, due to the limitation of the expressive ability of the SH lighting model, \cite{Kanamori,Manuel,tajimaPG21} could only produce low-frequency shadows and their method may fail under harsh illuminations.

\subsection{Inverse Rendering}

Ramachandran \textit{et al.}\cite{ramachandran1988perception} proposed the \textit{shape-from-shading} method to estimate shape from shading given an input image with known lighting conditions. Subsequent works \cite{christou1997light,okatani1997shape,lopez2009light} assume a simple light source model such as directional and point light sources. Based on the Retinex theory \cite{land1971lightness}, intrinsic images \cite{barron2012color,ye2014intrinsic,bonneel2017intrinsic,baslamisli2018cnn,laffont2015intrinsic,sheng2018intrinsic,li2018physically,ding2017intrinsic} aims to decompose an input image into reflectance and shading. Recent single-image human relighting studies \cite{Kanamori,Manuel,tajimaPG21} have drawn on the idea of intrinsic images to decompose an input image into albedo, shape and illumination using UNet. We have further improved the intrinsic images decomposition results in the de-lighting stage utilizing the modified HRNet \cite{sun2019deep} and achieved better disentanglement between albedo and lighting. 

% \vspace{-0.2cm}
\section{Overview}

\begin{figure*}[t]
\centering
\setlength{\abovecaptionskip}{0.2cm}
% \fbox{\rule{0pt}{2in} \rule{.9\linewidth}{0pt}}
\includegraphics[width=0.99\textwidth]{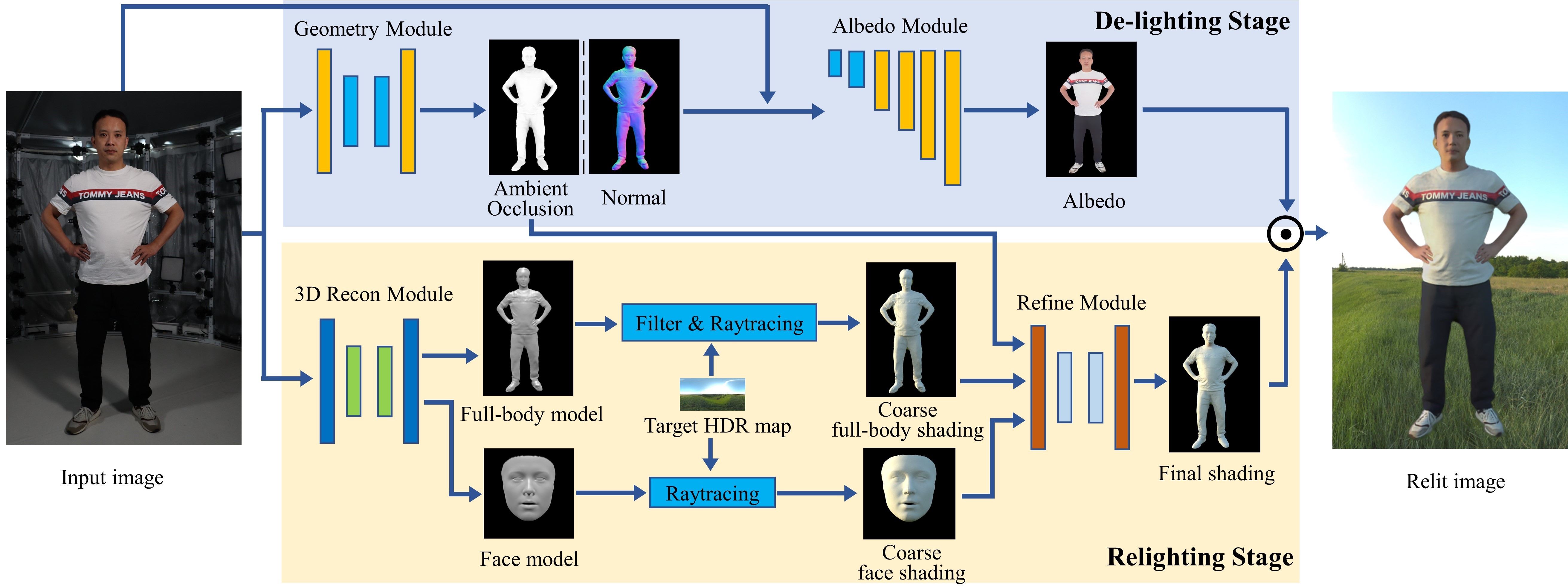} 
\caption{Illustration of our framework architecture. There are two stages in our method: de-lighting and relighting. The de-lighting stage takes the input image and outputs estimated albedo $\widehat{A}$ (Section 4). For the relighting stage (Section 5), a full-body 3D model and a face 3D model are estimated by 3D Recon Module and then are sent to the renderer to render coarse shading maps (Section 5.1). The Refine Module takes the coarse shading maps and the inferred ambient occlusion map as input and produces the final shading map (Section 5.2).}
% Finally, the estimated albedo map and final shading map are dot product to produce the relit image.}
\label{fig:framework}
% \vspace{-0.4cm}
\end{figure*}

Following current work \cite{wang2020single}, our framework consists of two stages: de-lighting and relighting. Fig.~\ref{fig:framework} shows the architecture of the whole framework. 
For the de-lighting stage, we first use two geometry networks (Geometry Module) to infer the per-pixel normal map $\widehat{N}$ and ambient occlusion map $\widehat{AO}$ separately. 
% The inferred normal map and ambient occlusion map are then used as auxiliary input to provide surface geometry and self-occlusion information for the de-lighting network and shading refinement network, respectively. 
Similar to \cite{pandey2021total}, the inferred normal map and the input image are concatenated as the input of Albedo Module to infer albedo $\widehat{A}$. 
% We modify the HRNet \cite{sun2019deep} as our de-lighting network (Albedo Module).
% by removing downsampling operations, avoiding skip connections and fusing multiscale features. 
For the relighting stage, we first render the 3D human models estimated by \cite{saito2020pifuhd} and \cite{guo2020towards} (3D Recon Module) using ray tracing to obtain the coarse full-body shading map $S_{coarse}^{body}$ and coarse face shading map $S_{coarse}^{face}$. Then the Refine Module takes these coarse shading maps and the inferred ambient occlusion map as input to produce the refined full-body shading map $\widehat{S_{fine}^{body}}$ and refined face shading map $\widehat{S_{fine}^{face}}$, which are composited together to obtain the final shading map $\widehat{S}$. Details regarding the specific architecture and implementations of the networks are provided in the supplementary materials. Finally, the dot product of the inferred albedo and the final shading map is obtained to produce the relit image $I_{relit}$ using the following formula:
 \[ I_{relit}=\widehat{A} \odot \widehat{S}\]

% \vspace{-0.2cm}
\section{De-lighting}
One of the most important factor for achieving better de-lighting results is the network architecture. 
Although the UNet architecture has been widely used in previous de-lighting networks~\cite{Manuel,Kanamori,pandey2021total,wang2020single,sengupta2018sfsnet}, we find that it usually produces heavy lighting and albedo entangled results, possibly because of the fact that the lighting and albedo features remain coupled in the UNet architecture.

\begin{figure}[t]
\centering
\setlength{\abovecaptionskip}{0.cm}
\includegraphics[width=0.98\textwidth]{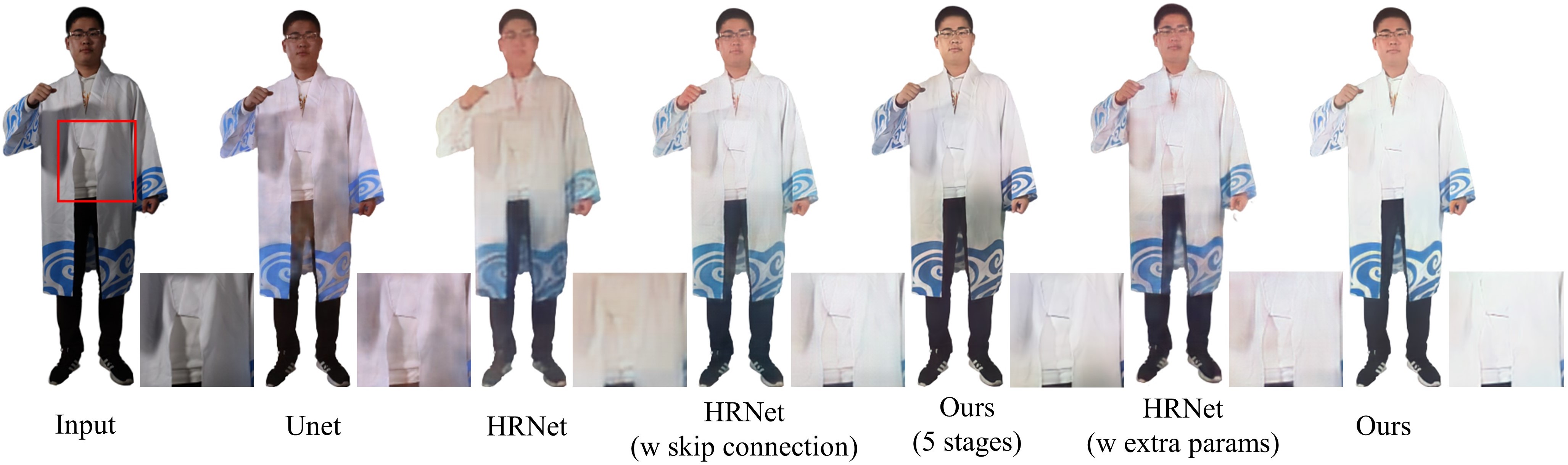}
\caption{Estimated albedo of a real image. Unet comes from RH \cite{Kanamori} and is retrained on our dataset. We zoom in the area in the red anchor and place it at the right of corresponding image.}
\label{fig:albedo_ablation}
% \vspace{-0.4cm}
\end{figure}

To guarantee that the lighting and albedo can be disentangled in the network architecture level, we modified the HRNet~\cite{sun2019deep} architecture and achieved better delighting performance than UNet as shown in Fig.~\ref{fig:albedo_ablation}. 
Note that the vanilla HRNet can only produce oversmoothed albedo map due to the aggressive downsampling operations in the beginning stage. 
Even with skip connections between the input downsampling features and output upsampling features, the results of vanilla HRNet remain poor and even make unremoved shadows appear in the inferred albedo map, especially in outdoor scenarios. 
We assume that skip connections have a negative impact on the final de-lighting results because high-resolution features from the input still contain environmental lighting information. Based on this assumption, we modify the vanilla HRNet by removing the downsampling operations directly (which also avoid skip connections) to fuse multiscale features while maintaining high-resolution representations. 
Moreover, we modify HRNet to output lighting prediction at the layer with the lowest resolution of the final stage. This strategy guarantees a better decomposition between lighting and albedo at both the architecture and feature representation level. 
 %To be more specific, we change the convolution stride of stage1 to 1 and extend HRnet to 6 stages. 
 %The last five stages number modules is [1,2,2,2,2]. 
Details regarding the specific architecture and implementations are provided in the supplementary materials.

To estimate the albedo of an input human image, we employ losses as follows:
\begin{equation}
    \begin{aligned}
        L_{DL}&= \lambda_{input}\left \| \widehat{A}-A \right \|_{1} +\lambda_{vgg} Vgg(\widehat{A},A) \\ &+\lambda_{light}\left \| w\odot log(1+\widehat{I_l})-log(1+I_l) \right \|_{2}^2 \nonumber
    \end{aligned}
\end{equation}
where $\widehat{A}$ is the inferred albedo and $A$ is the ground truth albedo. 
Similar to portrait relighting \cite{sun2019single}, %we modify the HRNet to output lighting prediction at the layer with the lowest resolution of the final stage and
we use the weighted log-$L_2$ loss. 
$\widehat{I_l}$ and $I_l$ are latitude-longitude representation of environmental illumination and are used to describe the estimated lighting and ground truth lighting, respectively. The ground truth lighting map is downsampled to $16\times32\times3$ using Gaussian pyramid. $w$ is the solid angle of each “pixel”. The $L_{DL}$ is the total loss for the de-lighting stage, and $\lambda_{input}$, $\lambda_{vgg}$ and $\lambda_{light}$ are the weight factors. Empirically, we find $\lambda_{input}=500$, $\lambda_{vgg}=100$, $\lambda_{light}=0.025$ achieves the best performance. 

% \vspace{-0.2cm}
\section{Relighting}

To render photorealistic shadows, we divide the relighting stage into two substages: ray tracing and shading refinement. We first estimate 3D models of the target human because the classical ray tracing algorithm requires a 3D model of target object.

For geometry estimation, given a single human image, we use PIFuHD \cite{saito2020pifuhd} to estimate a complete 3D human model. Then we crop the human face from the input image and align it with the dense 3D face model following \cite{guo2020towards} by regressing the 3DMM parameters. Both the full-body 3D model and face 3D model have no texture because we only need to render shading maps of the target human. 

\subsection{Ray Tracing}
For photorealistic rendering, we use the Cycles rendering engine in Blender \cite{blender.org} and a Principled BSDF shader. We set the camera mode to orthographic to ensure that the rendered shading map is pixel-aligned with the input image.  Due to the limitations of the PIFuHD \cite{saito2020pifuhd} network's generation capacity and memory space, the surface of the estimated full-body 3D model is not smooth enough, which may produce checkered artifacts after raytracing. The schematic diagram of the artifacts is provided in the supplementary materials.
% as shown in Fig.~\ref{fig:raytracing_abalation}. 
To solve the problem, we adopt Laplacian smoothing \cite{sorkine2005laplacian} for the full-body 3D model and set smoothing steps to 10 with cotangent weight. Then the smoothed full-body model is sent to the renderer to render the full-body shading map under the target lighting environment using ray tracing. In addition, the generated 3D face model was directly sent to the renderer without smoothing for ray tracing.

In a word, we render a coarse full-body shading map $S_{coarse}^{body}$ and coarse face shading map $S_{coarse}^{face}$ under the target lighting environment using ray tracing. The rendered coarse shading maps are pixel-aligned with the input image.

\subsection{Shading Refinement}

After ray tracing under the target lighting environment, we obtained roughly correct shadows (especially for self-occlusions and hard shadows such as cast shadows) of the target human. 
However, the full-body 3D model reconstructed from a single image is not completely accurate and thus the ray-traced shading maps contain unnatural shadows and obvious geometry errors. 
% This is because that the ray-tracing performance is heavily dependent on the accuracy of the 3D surface geometry. 
To enhance the realism of the ray-traced full-body shading map, we introduce two refinement networks to compensate for shadow-rendering errors and restore a high-quality shading map. 
The first refinement network is designed to paint in the ray-traced full-body shading map and improve the overall quality of the shading details. 
The second refinement network is designed to refine facial shading details to ensure that we can fully leverage the geometry priors of human faces. Fig.~\ref{fig:refine_module} shows the entire shading refinement process. 

% \vspace{-0.2cm}
\subsubsection{Full-body Shading Refinement}
\begin{figure}[t]
\centering
\setlength{\abovecaptionskip}{0.cm}
\includegraphics[width=0.98\textwidth]{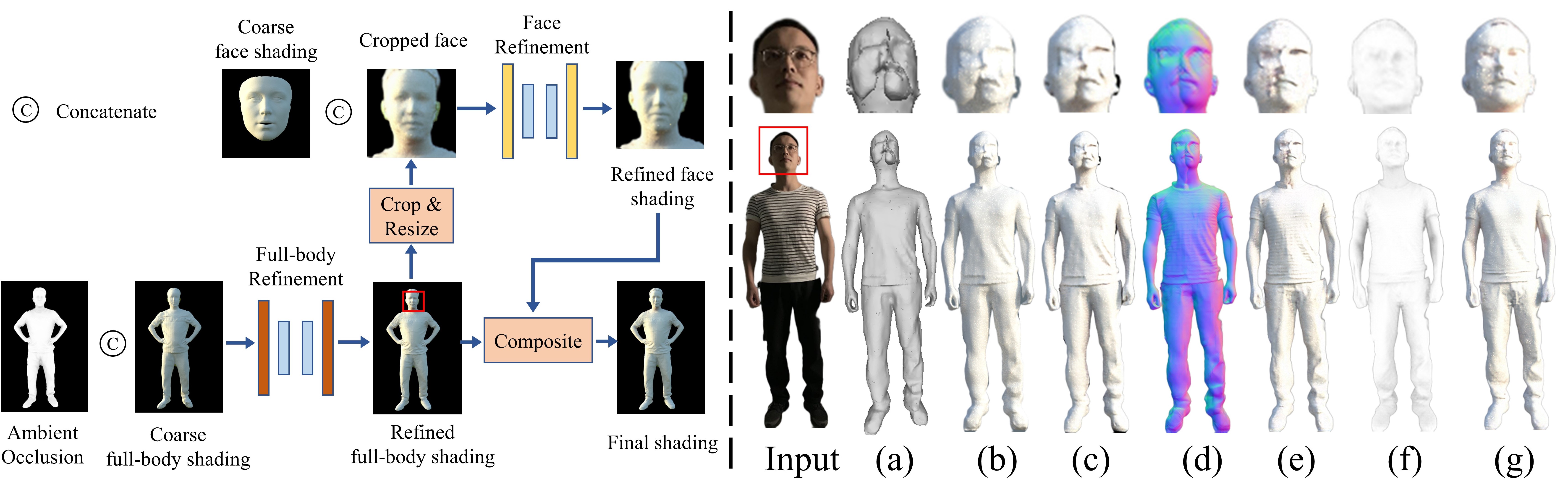}
\caption{\textbf{Left}: Illustration of Refine Module. A refined full-body shading map is inferred by the full-body refinement network. Then the cropped face from the refined full-body shading map and coarse face shading map is concatenated as the input of face refinement network, which outputs the refined face shading map. Finally, the refined face shading map and the refined full-body shading map are composited to generate the final relit shading map. \textbf{Right}: Refined shading maps.
% using the inferred normal map and ambient occlusion map as input, respectively. 
(a) 3D model estimated by PIFuHD \cite{saito2020pifuhd} (b) Ray-traced shading map (c) Refined shading map without the inferred ambient occlusion map (d) Inferred normal map (e) Refined shading map with the inferred normal map (f) Inferred ambient occlusion map (g) Refined shading map with the inferred ambient occlusion map. Cropped faces of corresponding shading map are placed at the top.}
\label{fig:refine_module}
% \vspace{-0.6cm}
\end{figure}
The full-body refinement network takes the coarse full-body shading $S_{coarse}^{body}$ and the inferred ambient occlusion map $\widehat{AO}$ as input and outputs the refined full-body shading residual. 
% It can be described as:
% \[\widehat{S_{fine}^{body}}= S_{coarse}^{body}+f_{refinebody}(S_{coarse}^{body},\widehat{AO}) \]
% where $f_{refinebody}$ is the full-body refinement network and $\widehat{S_{fine}^{body}}$ is the refined full-body shading map. 
% We draws on the idea of image restoration to use inferred ambient occlusion map as auxiliary input, so that the network can eliminate geometric errors of coarse shading map utilizing extra geometric information. 
We choose an ambient occlusion map instead of a normal map for the following three reasons. 
First, the ambient occlusion map can supplement part of the self-shadows lost due to geometry prediction errors. 
Second, compared with inferring a normal map, inferring an ambient occlusion map is more robust under various lighting environments even with extreme lighting distributions. 
Third, existing 3D human reconstruction  methods such as \cite{saito2020pifuhd,jafarian2021learning} are highly dependent on the surface normal to predict 3D geometry surface details, which means that the normal prediction errors are consistent with the geometry errors and $S_{coarse}^{body}$ cannot obtain extra correct geometry information from the normal map to compensate for existing shading errors.
Fig.~\ref{fig:refine_module} shows the refined results with the normal and ambient occlusion maps as auxiliary inputs, respectively.

The architecture of full-body refinement network is similar to MIMO-UNet \cite{cho2021rethinking}, and we deepen the network to 4 downsampling operations. The coarse full-body shading map $S_{coarse}^{body}$ and the inferred ambient occlusion map $\widehat{AO}$ are concatenated together as the original scale input and the downsampled $\widehat{AO}$ is used as multiscale input. We use multiscale output for supervision. We add adversarial loss in the final highest-resolution output layer to help the network generate plausible shading effects and use PatchGAN as the discriminator.
The training loss consists of content loss, fft loss and PatchGAN loss and is defined as follows:
\begin{equation}
\setlength\abovedisplayskip{0pt}
\setlength\belowdisplayskip{0pt}
    \begin{aligned}
        L_{fb} & =  \lambda_{content}\sum_{k=1}^{K}\left \| \widehat{S}_{fine}^{body^k}-S^k \right\|_{1}+\left \| P(\widehat{S}_{fine}^{body})-P(S) \right\|_{2}^2 \\ &+\lambda_{fft}\sum_{k=1}^{K}\left \| F(\widehat{S}_{fine}^{body^k})-F(S^k) \right \|_{1} \nonumber
    \end{aligned}
\end{equation}
where $K$ is the number of levels and $K=5$; $\widehat{S}_{fine}^{body^k}$ is the $k_{th}$ level output and $S^k$ is the $k_{th}$ downsampled ground truth shading map. $P$ is the PatchGAN discriminator, and $F()$ denotes the fast Fourier transform (FFT).
% which is designed to restore high-frequency details.
$\lambda_{content}$ and $\lambda_{fft}$ are weight factors and we empirically set $\lambda_{content}=10$ and $\lambda_{fft}$=0.001.

% \vspace{-0.2cm}
\subsubsection{Face Shading Refinement}
Although the above-mentioned refinement network produces realistic shading with rich details, the human eye is very sensitive to the details of the face and is able to distinguish small geometry and shadow errors. Therefore, we continue to refine the face region on the basis of the refined full-body shading map $\widehat{S_{fine}^{body}}$. The face refinement network takes $S_{crop}^{face}$ cropped from $\widehat{S_{fine}^{body}}$ and $S_{coarse}^{face}$ as the input and outputs the refined face shading residual.  The training loss can be expressed as follows:
\[ L_{ff}=\lambda_{face}\left \| \widehat{S_{fine}^{face}}-S^{face} \right \|_{1}+\left \|L(\widehat{S_{fine}^{face}})-L(S^{face}) \right\|_{2}^2\]
where $L$ is the LSGAN discriminator, $\lambda_{face}$ is the weight factor and is set to be $\lambda_{face}=5$. $S^{face}$ is the ground truth face shading map. We use UNet architecture as the backbone of face refinement network and details regarding the architecture are provided in the supplementary materials.

% \vspace{-0.3cm}
\section{Implementation Details}

We carefully select 811 scanned 3D human figures with good lighting conditions from Twindom \cite{twindom}, of which 700 figures are used for training and 111 figures for testing. The use of the dataset has been officially approved by Twindom. We collect 480 panoramic lighting environments sourced from www.HDRIHaven.com \cite{hdr} and rotate them every 36 degrees to generate a total of 4800 HDR environment maps. We allocate 4600 HDR environment maps for training and 200 HDR environment maps for testing. To balance the amount of lighting in indoor and outdoor scenes, we add an extra 150 indoor HDR environment maps from the Laval Indoor Dataset \cite{gardner2017learning} to the testing dataset. None of the test lighting conditions appear in the training dataset. 
% We use the Cycle rendering engine in Blender \cite{blender.org} and a Principled BSDF shader to render datasets. 
Details regarding the specific data rendering, training and testing are provided in the supplementary materials. 

% \vspace{-0.2cm}
\section{Experiments}

\begin{figure*}[t]
\centering
% \fbox{\rule{0pt}{2in} \rule{.9\linewidth}{0pt}}
\setlength{\abovecaptionskip}{0.cm}
\includegraphics[width=0.99\textwidth]{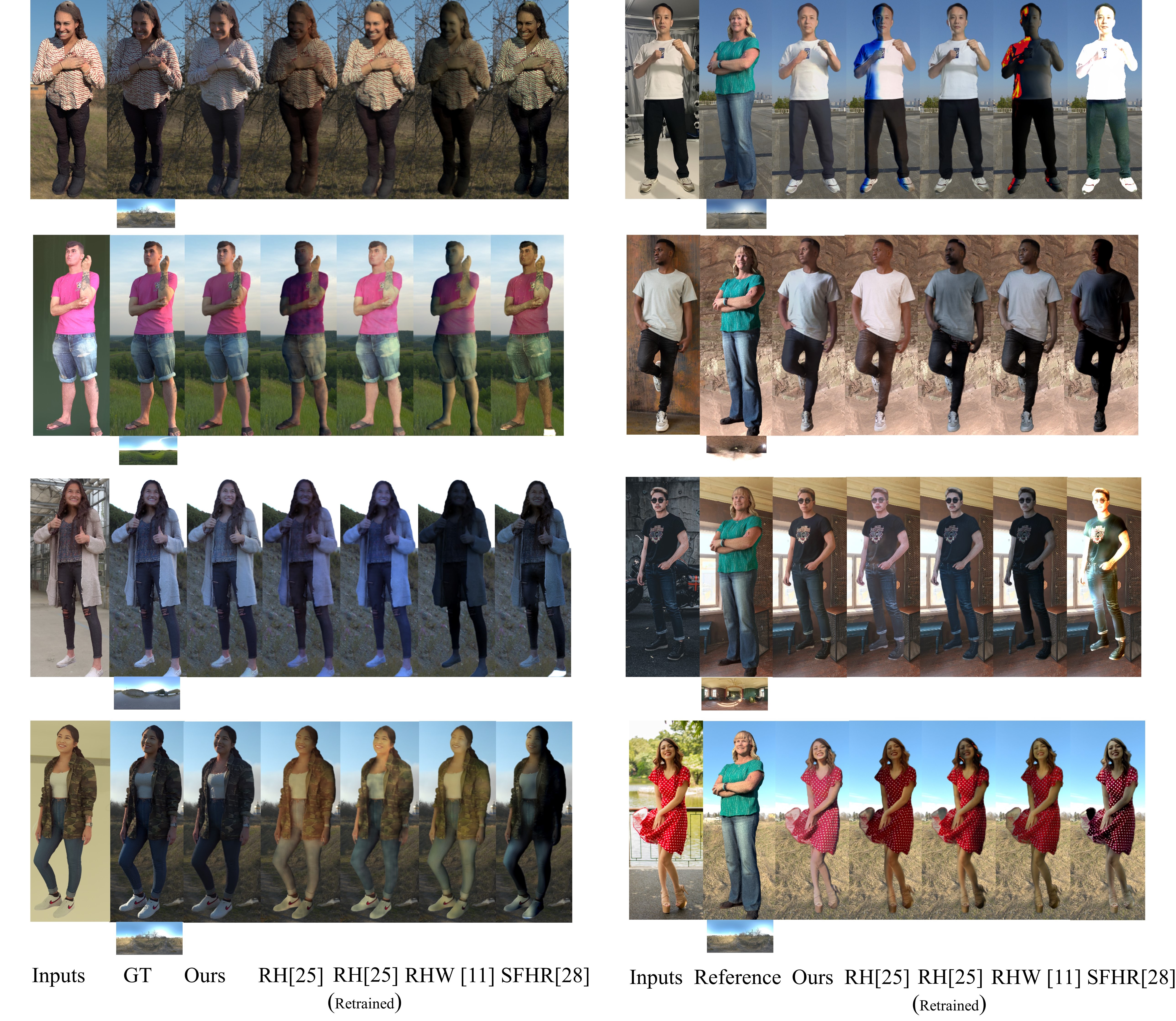} 
\caption{Relit results on synthetic and real images. The first column: synthetic images from our testing dataset. The second column: real images. For real images, ''Reference'' are the rendered images of a virtual 3D human model under the target lighting conditions and are used to indicate the position of shadows. The target HDR environment map is placed under the ground truth image or reference image.}
% Note that our method can not only remove the hard shadows of the input image, but also generate plausible hard shadows in the relit image.}
\label{fig:comp_real_relight}
% \vspace{-0.6cm}
\end{figure*}

In this section, we first compare our method with previous state-of-the-art methods quantitatively and qualitatively to show that our method performs better on de-lighting task and produces more photorealistic relit results under challenging lighting conditions. Then we evaluate the key contributions of our proposed method and prove the effectiveness of the entire framework and each module. For quantitative comparisons, we adopt the metrics MSE, PSNR and SSIM to compare the inferred albedo, relit shading and relit images with the corresponding ground truth images in our testing dataset. For
qualitative comparisons, we show some intrinsic images decomposition results and relit results under target lighting conditions using both real images and synthetic images. 

\subsection{Comparisons with SOTA Methods}

We compare our method with the state-of-the-art single-image human relighting methods RH \cite{Kanamori}, SFHR \cite{Manuel} and RHW \cite{tajimaPG21}. 
All of them are based on intrinsic images decomposition and require only a single human image and target lighting for relighting as in our approach. Details regarding the specific comparison settings are provided in the supplementary materials. 

\begin{table}[t]
\small
\caption{Quantitative comparisons of our single-image human relighting framework against prior works. Shading indicates the estimated shading map under target lighting condition. The value of MSE is scaled by multiplying $10^{3}$.}
\begin{center}
\setlength{\tabcolsep}{1pt}{
\begin{tabular}{c|ccc|ccc|ccc}
\hline
\multirow{3}{*}{} & \multicolumn{3}{c|}{Relight} & \multicolumn{3}{c|}{Albedo} & \multicolumn{3}{c}{Shading} \\
                  & MSE $\downarrow$     & SSIM $\uparrow$     & PSNR $\uparrow$    & MSE $\downarrow$     & SSIM $\uparrow$    & PSNR $\uparrow$    & MSE $\downarrow$     & SSIM $\uparrow$    & PSNR $\uparrow$    \\ \hline
RH \cite{Kanamori}                &   8.501      &  0.9275        & 23.70        &    4.528     &   0.9521      &   24.69      &    18.96     &    0.9162     &    20.26     \\
\begin{tabular}[c]{@{}c@{}}RH \cite{Kanamori}\\ (Re-trained)\end{tabular}  &     3.781    &  0.9506        &    26.45     &  2.819       &  0.9559       &   26.73      &    8.333     &  0.9236       &  22.39       \\
RHW \cite{tajimaPG21}              &     10.90    &  0.9085        &     22.97    &  3.624       &  0.9586       &  25.28       & 26.18        &    0.9037     &    19.43     \\
SFHR \cite{Manuel}             &     5.833    &  0.9349        & 24.66        &  4.185       &  0.9613       &  24.98       &    12.79     &     0.9155    &     21.41    \\
Ours              &   \textbf{1.311}      &     \textbf{0.9638}     &   \textbf{30.20}      &   \textbf{1.468}      &  \textbf{0.9814}       &  \textbf{30.27}       &     \textbf{0.729}    & \textbf{0.9626}        &    \textbf{32.86}    
\end{tabular}}
\end{center}
\label{tab:Quantitative_evaluations}
% \vspace{-0.8cm}
\end{table}

% \vspace{-0.3cm}
\subsubsection{Comparison on Synthetic Data}

We first perform quantitative and qualitative comparisons on the testing dataset where we have ground truth images as a comparison. Tab.~\ref{tab:Quantitative_evaluations} shows quantitative comparison of intrinsic images decomposition performance and single-image human relighting quality. Our method outperforms the competitive methods on every metric for both tasks. 
To limit the comparison to decomposition and relighting quality only, all metrics are computed only on the foreground region for all methods. 

Qualitative comparisons for relighting are shown in Fig.~\ref{fig:comp_real_relight}. 
To improve the visual effects, we use MODNet \cite{ke2020green} to infer alpha channel of the input image and change the background of the relit images to the corresponding part of the environment lighting map. 
Benefiting from the combination of classical forward rendering and deep learning, our method can produce photorealistic shadows under arbitrary lighting conditions. 
As a comparison, RH \cite{Kanamori}, SFHR \cite{Manuel} and RHW \cite{tajimaPG21} can only produce low-frequency shadows. 
% thus lack realism. 
Moreover, they also fail to relight images under outdoor lighting conditions and produce overly bright or dark relit results. This is due to the fact that their models require target illuminations to have a reference radiance in the range of [0.7,0.9], which greatly restricts their performance and broad applications. 
As shown in Fig.~\ref{fig:shading_syns}, our method can produce plausible hard cast shadows and achieve better disentanglement between albedo and shading, which remain challenging for previous methods. 
Since all methods rely on albedo estimation at first for relighting, we also compare the de-lighting performance. 
As shown in Fig.~\ref{fig:delight_syns}, compared with RH \cite{Kanamori}, SFHR \cite{Manuel} and RHW \cite{tajimaPG21}, our method achieves better disentanglement between lighting and albedo and is able to remove large areas of shadows (even including self-shadows from concave regions on the real-world 3D shape of the human body, e.g., armpits, crotch, neck under the chin or folds on the clothing.
%Relightinghuman and FBHR seem to just adjust the brightness of the shadow and there are still most of the shadows left in albedo.

% \vspace{-0.2cm}
\subsubsection{Comparison on Real-world Images}

\begin{figure*}[t]
\centering
% \fbox{\rule{0pt}{2in} \rule{.9\linewidth}{0pt}}
\setlength{\abovecaptionskip}{0.cm}
\includegraphics[width=0.99\textwidth]{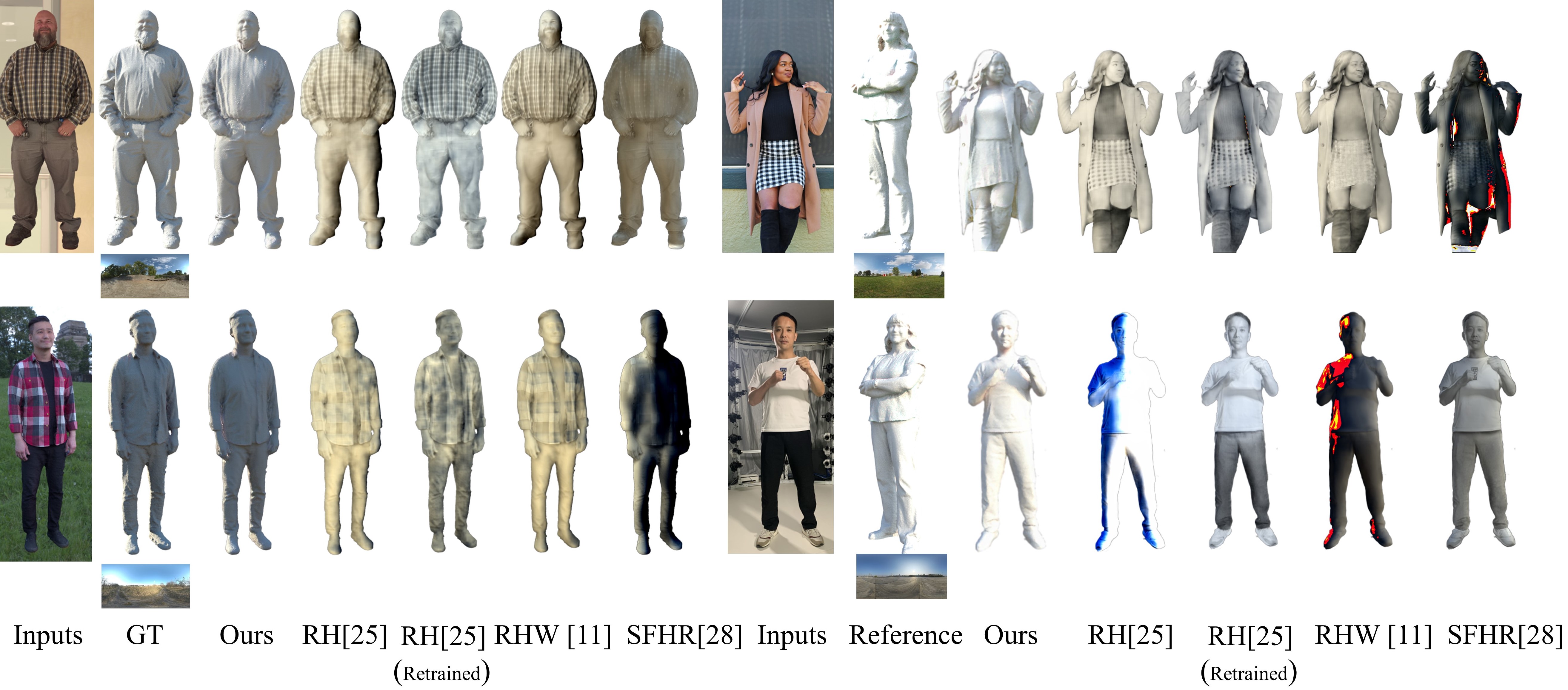} 
\caption{Qualitative results for shading estimation under target lighting conditions. The first column: synthetic images from our testing dataset. The second column: real images.}
% Note that our method can not only generate high-frequency shadows such as cast shadows, but also achieve better disentanglement between albedo and shading.}
\label{fig:shading_syns}
% \vspace{-0.2cm}
\end{figure*}

\begin{figure*}[t]
\centering
\setlength{\abovecaptionskip}{0.cm}
% \fbox{\rule{0pt}{2in} \rule{.9\linewidth}{0pt}}
\includegraphics[width=0.99\textwidth]{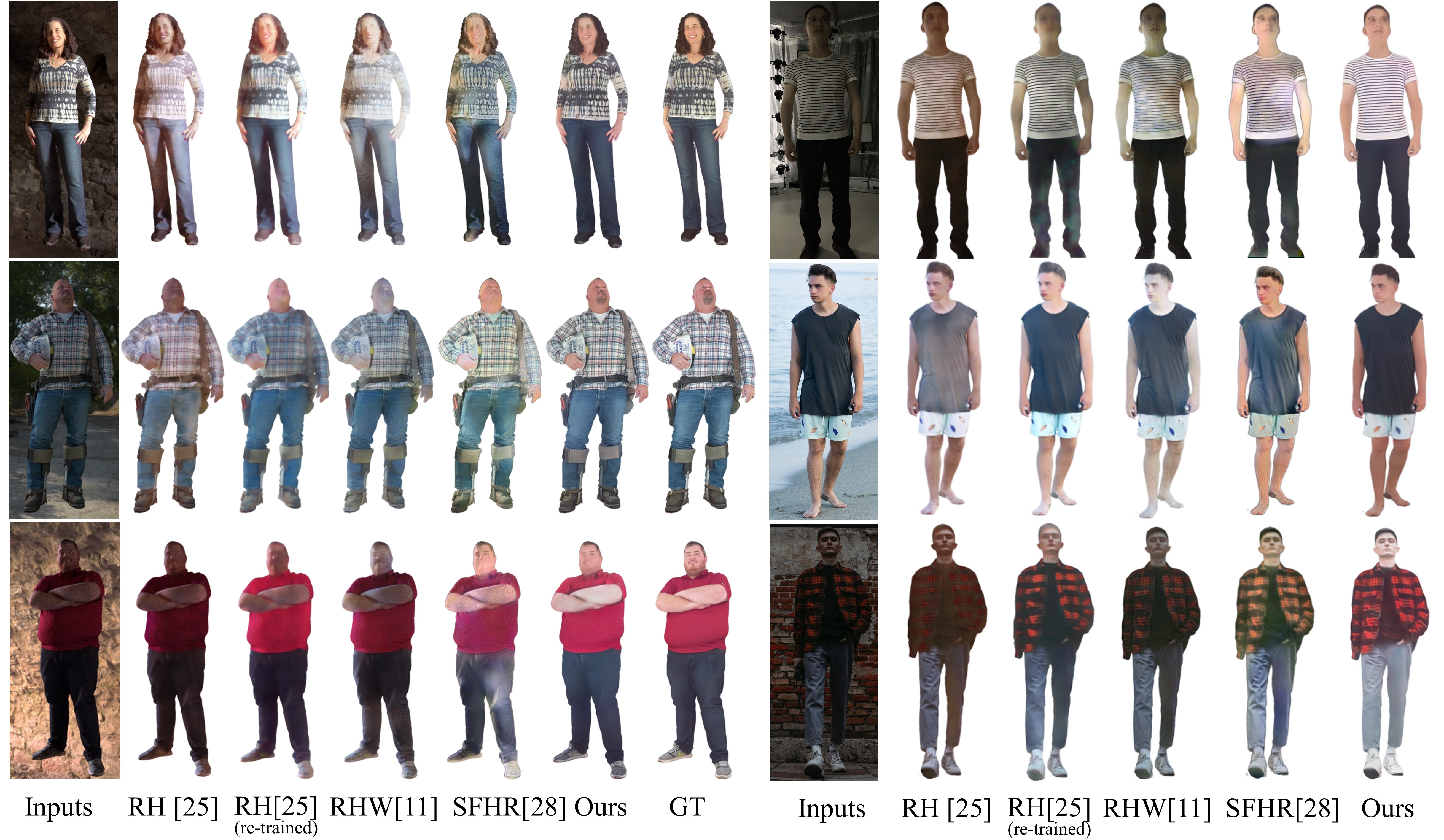} 
\caption{De-lighting results on synthetic and real images. The first column: synthetic images from our testing dataset. The second column: real-world images. }
\label{fig:delight_syns}
% \vspace{-0.2cm}
\end{figure*}

Although our method is trained on synthetic datasets, it is generalizable to real data. The second column of Fig.~\ref{fig:shading_syns} and Fig.~\ref{fig:delight_syns} show shading estimation and de-lighting results on real images, respectively, and the second column of Fig.~\ref{fig:comp_real_relight} shows the relit results of images photographed in the real world under arbitrary and complex illumination conditions. 
% Once again, our method outperforms other state-of-the-art methods in both de-lighting and relighting. 
RH \cite{Kanamori}, SFHR \cite{Manuel} and RHW \cite{tajimaPG21} still suffer from the entanglement of lighting and albedo and struggle to produce high-frequency shadows. 
By comparison, our method performs better at removing detailed shadows from the original images and generating photorealistic shadows on the relit images. %Please refer to the supplementary materials for more results.

\subsection{Ablation study}
To demonstrate the effectiveness of our delighting network, full-body refinement network and face refinement network, we conduct comprehensive ablation
studies both quantitatively and qualitatively. 

\begin{table}[t]
\caption{Quantitative results for ablation study of the full-body shading refinement.}
\begin{center}
\setlength{\tabcolsep}{3pt}{
\begin{tabular}{c|ccc}
\hline
                                                                             & MSE $\downarrow (\times 10^{-3})$ & SSIM $\uparrow$ & PSNR $\uparrow$ \\ \hline
Raytracing                                                                 &              1.492            &               0.9499            &   29.84   \\
Ours(w/o ambient)                                                                &            1.250              &              0.9524             &   30.89   \\
Ours(w normal) &            0.886                &     0.9561                      &   32.29    \\
Ours                                                                         &         \textbf{0.729}             &          \textbf{0.9626}               &     \textbf{32.86} 
\end{tabular}}
\end{center}

\label{tab:ablation_refinement}
% \vspace{-0.8cm}
\end{table}

\begin{table}[t]
\caption{Quantitative results for ablation study of the de-lighting network. Vanilla Unet comes from RH \cite{Kanamori} and is trained on our dataset.}
\begin{center}
\setlength{\tabcolsep}{3pt}{
\begin{tabular}{c|ccc}
\hline
                                                                             & MSE $\downarrow (\times 10^{-3})$ & SSIM $\uparrow$ & PSNR $\uparrow$ \\ \hline
Vanilla Unet                                                                 &              2.819            &               0.9559          &   26.73   \\
Vanilla HRNet                                                                &            2.470              &              0.9550             &   27.21   \\
\begin{tabular}[c]{@{}c@{}}Vanilla HRNet\\ (w skip connections)\end{tabular} &            2.150                &     0.9767                       &   28.53    \\
Ours(5 stages)                                                               &              1.728              &              0.9808             &   28.91   \\
Ours                                                                         &         \textbf{1.468}             &          \textbf{0.9814}               &     \textbf{30.27} 
\end{tabular}}
\end{center}
\label{tab:ablation_net_arch}
% \vspace{-0.6cm}
\end{table}
First, we compare our delighting network with vanilla UNet, vanilla HRNet and vanilla HRNet with skip connections. Tab.~\ref{tab:ablation_net_arch} shows the quantitative results of different networks and Fig.~\ref{fig:albedo_ablation} presents the de-lighting results on a real image.
% We also investigate the impact of the depth of the network architecture on delighting performance. 
The vanilla HRNet is HRNet-W32 \cite{sun2019deep} has two transposed convolution layers with stride 2 to ensure that the output size and input size are the same. 
Based on the vanilla HRNet, the vanilla HRNet with skip connections further adds skip connections between the downsampled features of the first stage and the transposed convolution features of the output. 
``HRNet(w extra params)'' means vanilla HRNet with skip connections, extra stages, modules and blocks. 
``Ours(5 stages)" indicates that the architecture of network is similar to that of our de-lighting network but only contains 5 stages.
% The experiments show that our 6-stage de-lighting network performs better in removing local shadows than a 5-stage de-lighting network, while the improvement brought by continuing to increase the depth of the network is marginal. Hence we finally chose to use a 6-stage network. 
The proposed network ``Ours'' outperforms other networks on all metrics and achieves better disentanglement between lighting and albedo. 
By contrast, the vanilla HRNet fails to produce high-resolution results and the removal of partial self-shadows such as those caused by clothing folds remains difficult, even with skip connections and extra parameters.
Compared with UNet, our method greatly improves PSNR (increasing by 3) and MSE (dropping to half of the UNet's). 

Second, we verify the effectiveness of the ambient occlusion map used by the full-body refinement network. 
``Refinement Net(w/o ambient)'' means that the refinement network only takes coarse full-body shading as input and removes SCM and FAM modules of MIMO-UNet\cite{cho2021rethinking}. 
``Refinement Net(w normal)'' means that the refinement network takes the normal map as the auxiliary input rather than the ambient occlusion map. 
Tab.~\ref{tab:ablation_refinement} shows quantitative results and Fig.~\ref{fig:refine_module} shows qualitative results. 
Without an ambient occlusion map, the refinement work cannot fill in missing geometry details and shadows. 
Moreover, when the input image is under extreme lighting conditions, the inference of normal map may fail around the boundary of the shadows. 
By contrast, the inferred ambient occlusion map is unaffected by shadows and able to recover more geometry and occlusion details, thus restoring better shading maps.

\begin{figure}[t]
\centering
\setlength{\abovecaptionskip}{0.0cm}
% \fbox{\rule{0pt}{2in} \rule{.9\linewidth}{0pt}}
\includegraphics[width=0.98\textwidth]{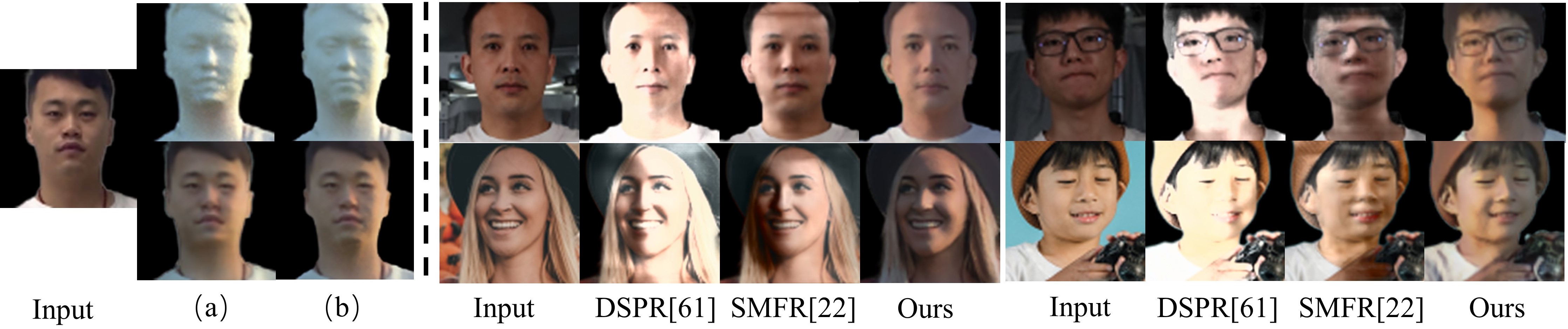}
\caption{\textbf{Left}: Ablation of face refinement module. (a) Top: cropped face from the refined full-body shading map Bottom: corresponding relit result (b) Top: refined face shading map Bottom: corresponding relit result. \textbf{Right}: Comparison with existing portrait relighting methods.}
\label{fig:face_refinement}
% \vspace{-0.6cm}
\end{figure}
Finally, we evaluate the face refinement network. To highlight the role of this module, we present the cropped face shading for comparison.
Fig.~\ref{fig:face_refinement} shows the qualitative results for face refinement. The face without refinement contains unnatural facial details such as a twisted nose and asymmetric eyes. By contrast, thanks to the geometry priors provided by 3DMM templates, the refined face owns clearer and more natural facial features. Compared with DSPR [61] and SMFR [22], our method can generate plausible hard cast shadows, especially around the nose and neck, whereas SMFR [22] may produces patchy shadows and DSPR [61] may produces overexposed results.

% \vspace{-0.2cm}
\section{Discussion}
\noindent
\textbf{Conclusion.} We propose a geometry-aware single-image human relighting framework that leverages 3D geometry prior information to produce higher-quality relit results. Our framework contains two stages: de-lighting and relighting. For the de-lighting stage, we use a modified HRNet as the de-lighting network and achieve better disentanglement between lighting and albedo. For the relighting stage, we use ray tracing to render the shading map of the target human, and further refine it utilizing learning-based refinement networks. The extensive results demonstrate that our framework can produce photorealistic high-frequency shadows with clear boundaries under challenging lighting conditions and outperforms the existing SOTA method on both synthetic images and real images. 

\noindent
\textbf{Limitations.} Due to the limitation of the dataset, we adopt the assumption of Lambertian materials for the clothed humans, which fails to produce specular reflectance in the relit results. For the same reason, our delighting network struggles to remove the highlights on the face. Moreover, inaccurate single-image geometry reconstruction may generate unnatural refined shading results. 

\noindent
\textbf{Acknowledgement. } This paper is supported by National Key R\&D Program of China (2021ZD0113501) and the NSFC project No.62125107, No.62171255 and No.61827805.

% \clearpage\mbox{}Page \thepage\ of the manuscript.
% \clearpage\mbox{}Page \thepage\ of the manuscript.

% This is the last page of the manuscript.
% Now we have reached the maximum size of the ECCV 2022 submission (excluding references).
% References should start immediately after the main text, but can continue on p.15 if needed.

\clearpage
\bibliographystyle{splncs04}
\bibliography{egbib}

\title{Geometry-aware Single-image Full-body Human Relighting Supplementary Materials} % Replace with your title

\author{Chaonan Ji\inst{1} \and Tao Yu\inst{1} \and
Kaiwen Guo\inst{2} \and Jingxin Liu\inst{3} \and Yebin Liu\inst{1}}
\authorrunning{C. Ji et al.}
\institute{Department of Automation, Tsinghua University, China \and
Meta Reality Labs \and Guangdong OPPO Mobile Telecommunications Corp., Ltd}

\titlerunning{Geometry-aware Single-image Full-body Human Relighting}
%\begin{comment}
% \titlerunning{ECCV-22 submission ID \ECCVSubNumber} 
% \authorrunning{ECCV-22 submission ID \ECCVSubNumber} 
% \author{Anonymous ECCV submission}
% \institute{Paper ID \ECCVSubNumber}
\maketitle
%\end{comment}
%******************

\section{Network Architecture}

% \begin{figure}[t]
% \centering
% % \fbox{\rule{0pt}{2in} \rule{0.9\linewidth}{0pt}}
%     \includegraphics[width=0.99\textwidth]{Images_supp/teaser.jpg} 
%     \caption{Given a single human image and an arbitrary high dynamic range lighting environment, our framework estimates a high-quality albedo map and generates photorealistic relit images of the target human subject under the desired lighting conditions.}
%   %\includegraphics[width=0.8\linewidth]{egfigure.eps}
% \end{figure}

\subsection{Geometry Module} 
Given an input image $I$, our geometry module performs image-to-image translation of the input image to its corresponding normal map $\widehat{N}$ and ambient occlusion map $\widehat{AO}$ using a U-Net architecture similar to pix2pixHD \cite{wang2018high}. We empirically found that training geometry networks separately produces more accurate results than joint training. 
 The losses for geometry estimation can be expressed as follows:
 \[ L_N=\lambda_{geo}{\left \| \widehat{N}-N \right \|_{1}}+Vgg(\widehat{N},N)\]
  \[ L_{AO}=\lambda_{geo}{\left \| \widehat{AO}-AO \right \|_{1}}+Vgg(\widehat{AO},AO)\]
 where $L_N$ is the normal loss, $L_{AO}$ is the ambient occlusion loss and $Vgg$ is the Vgg loss. $AO$ is the ground truth of ambient occlusion map and $N$ is the ground truth of normal map. $\lambda_{geo}$ is the weight factor and $\lambda_{geo}=5$.
 Note that we add skip connections for the ambient occlusion map inference network to restore more details.
 
 \subsection{De-lighting Network}
 \begin{figure}[t]
\centering
\includegraphics[width=0.90\textwidth]{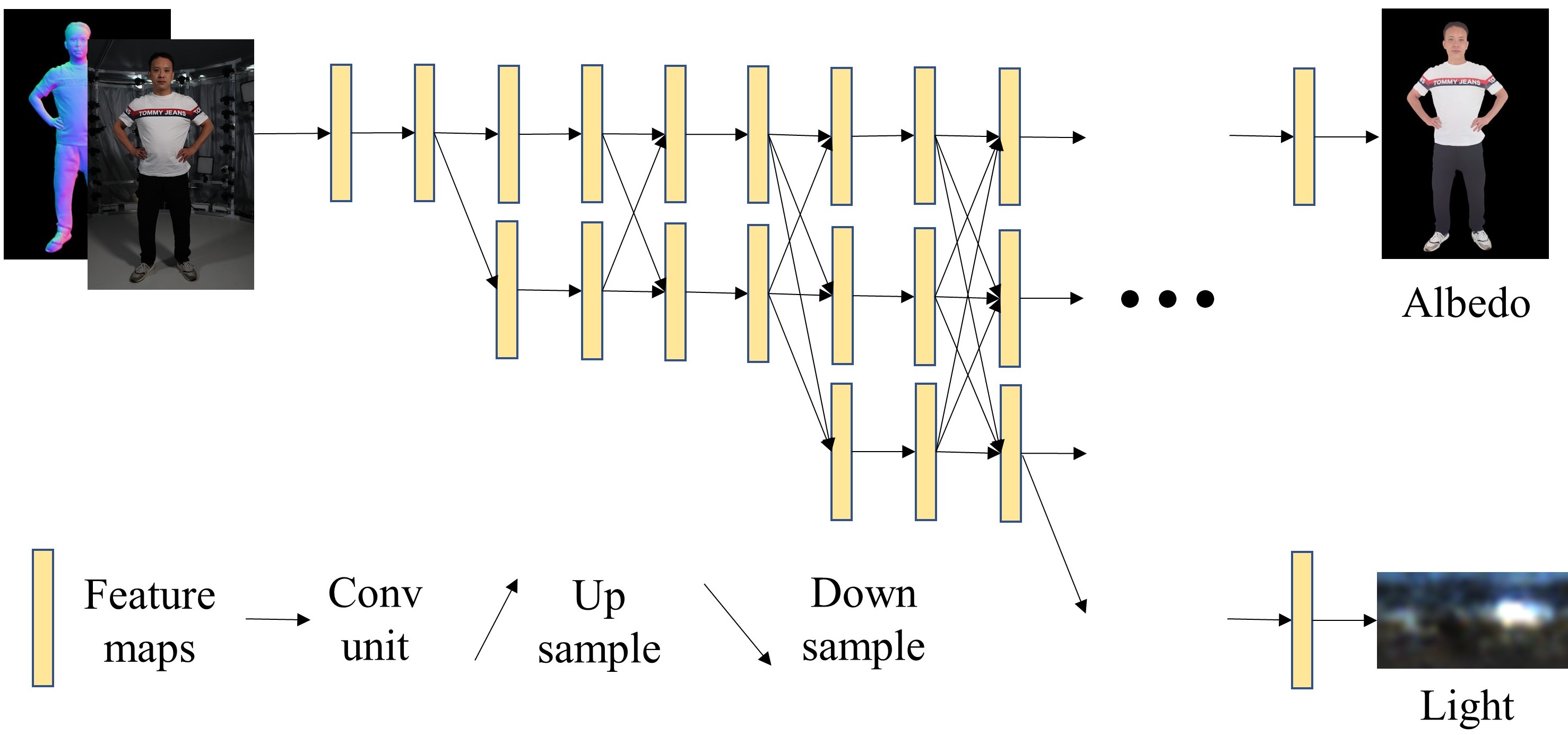} 
\caption{The architecture of the de-lighting network. The de-lighting network takes the input image and inferred normal map as input, and outputs albedo at the layer with the highest resolution of the final stage and light at the layer with the lowest resolution of the final stage. }
\label{fig:albedo_module}
% \vspace{-0.3cm}
\end{figure}

Fig.~\ref{fig:albedo_module} shows the architecture of the de-lighting network. We remove the the downsampling operations of the first stage in HRNet \cite{sun2019deep} directly (which also avoid skip connections) to fuse multiscale features while maintaining high-resolution representations. Our de-lighting network contains 6 stages and the number modules of last five stages are [1,2,2,2,2]. The ``NUM BLOCKS'' is set to 2 for the last five stages and the last layer is followed by an extra $1\times1$ convolution layer to generate output. The convolution stride of the first stage is set to 1 and the upsample mode is set to bilinear. ``Ours(5 stages)'' only contains 5 stages and the rest of the network is the same as ``Ours''. ``HRNet(w extra parameters)'' is adpated from HRNet-W32 \cite{sun2019deep} and owns 6 stages (the number modules of last five stages are [1,4,3,2,1]). The ``NUM BLOCKS'' is set to 4.

\subsection{Full-body Refinement Network}
 \begin{figure}[t]
\centering
\includegraphics[width=0.90\textwidth]{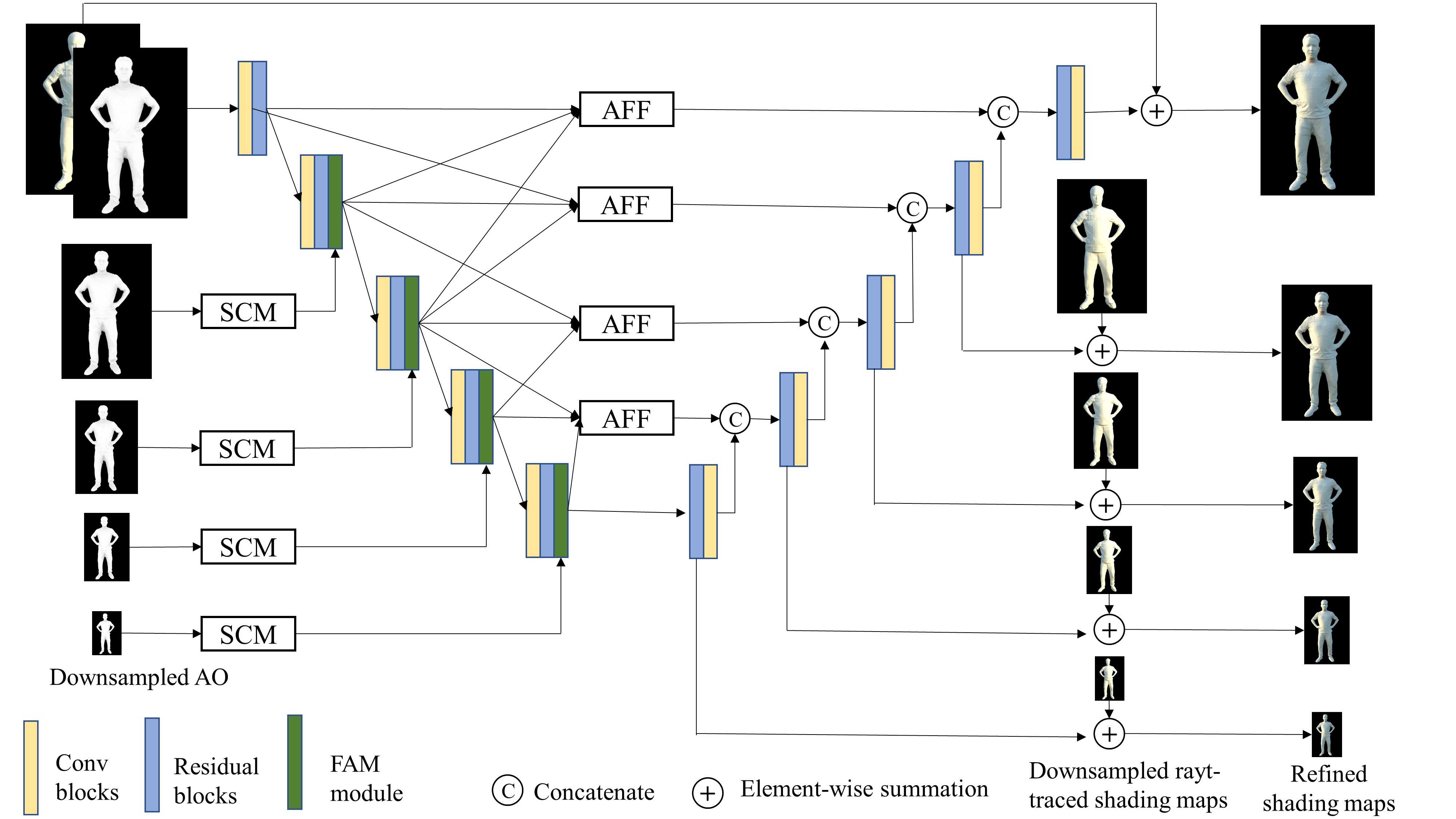} 
\caption{The architecture of the full-body shading refinement network.}
\label{fig:full_body_shading}
\end{figure}

Fig.~\ref{fig:full_body_shading} shows the architecture of the full-body shading refinement network. The network is modified from MIMO-Unet \cite{cho2021rethinking} and is deepen to 4 downsampling operations. It takes multi-scale input images as input and outputs multi-scale refined full-body shading maps. We concatenate the ray-traced shading map $S_{coarse}^{body}$ and inferred ambient occlusion map $\widehat{AO}$ as the input to the highest-resolution convolution layer and the downsampled ambient occlusion maps are used as the inputs to  low-resolution convolution layers. The FAM, SCM and AFF modules are the same with the MIMO-UNet \cite{cho2021rethinking}.

\subsection{Face Refinement Network}
 \begin{figure}[t]
\centering
\includegraphics[width=0.90\textwidth]{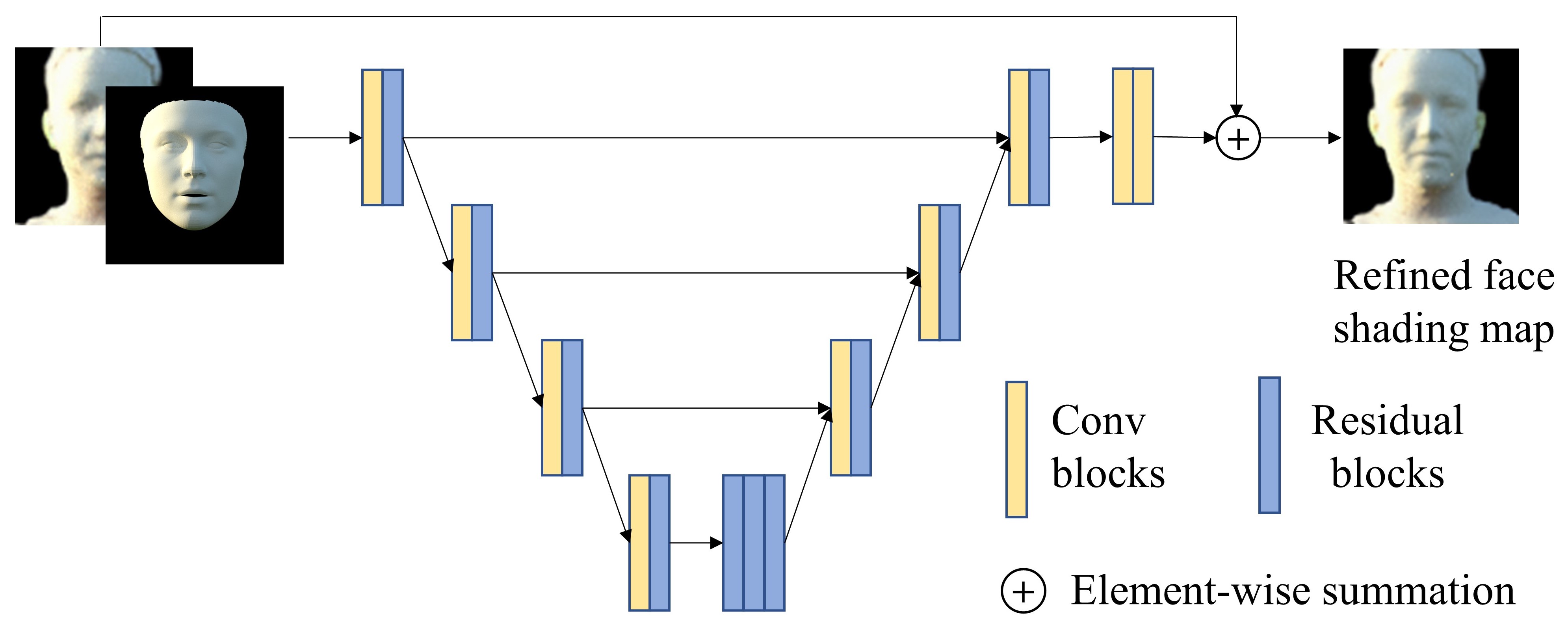} 
\caption{The architecture of the face shading refinement network.}
\label{fig:face_shading}
\end{figure}
Fig.~\ref{fig:face_shading} shows the architecture of the face shading refinement network. The network takes the cropped face from the refined full-body shading map $\widehat{S_{fine}^{body}}$ and ray-traced face shading map $S_{coarse}^{face}$ as input, and outputs the refined face shading map. The input is resized to $128 \times 128$. The network has 8 encoder-decoder layers and skip connections and each layer is run through $3\times3$ convolutions followed by LeakyReLU activations and BatchNorm ($7\times7$ convolutions for the first and last layer, and the last layer is followed by an extra $3\times3$ convolution layer to generate output). The number of filters are 64,128,256,512 for the encoder,512 for the bottleneck, and 256,128,64,64 for the decoder respectively. Each convolution layer except the last layer is followed by a residual block and there are 3 residual blocks in the bottleneck.

% \section{Implementation Details}
% \begin{figure*}[t]
% % \fbox{\rule{0pt}{2in} \rule{0.9\linewidth}{0pt}}
% \centering
% % \setlength{\abovecaptionskip}{0.cm}
% \includegraphics[width=0.99\textwidth]{Images_supp/shading.jpg}

% \caption{Qualitative results for shading estimation under target lighting conditions. The first column: synthetic images from our testing dataset. The second column: real images. For real images, ''Reference'' are the rendered shading maps of a virtual 3D human model under the target lighting conditions and are used to indicate the position of shadows. The target HDR environment map is placed under the ground truth image or reference image. Note that our method can not only generate high-frequency shadows such as cast shadows, but also achieve better disentanglement between albedo and shading. }
% \label{fig:shading}
% \vspace{-0.3cm}
% \end{figure*}

\section{Implementation Details}
\subsection{Data}
\begin{figure}[t]
\centering
\includegraphics[width=0.60\textwidth]{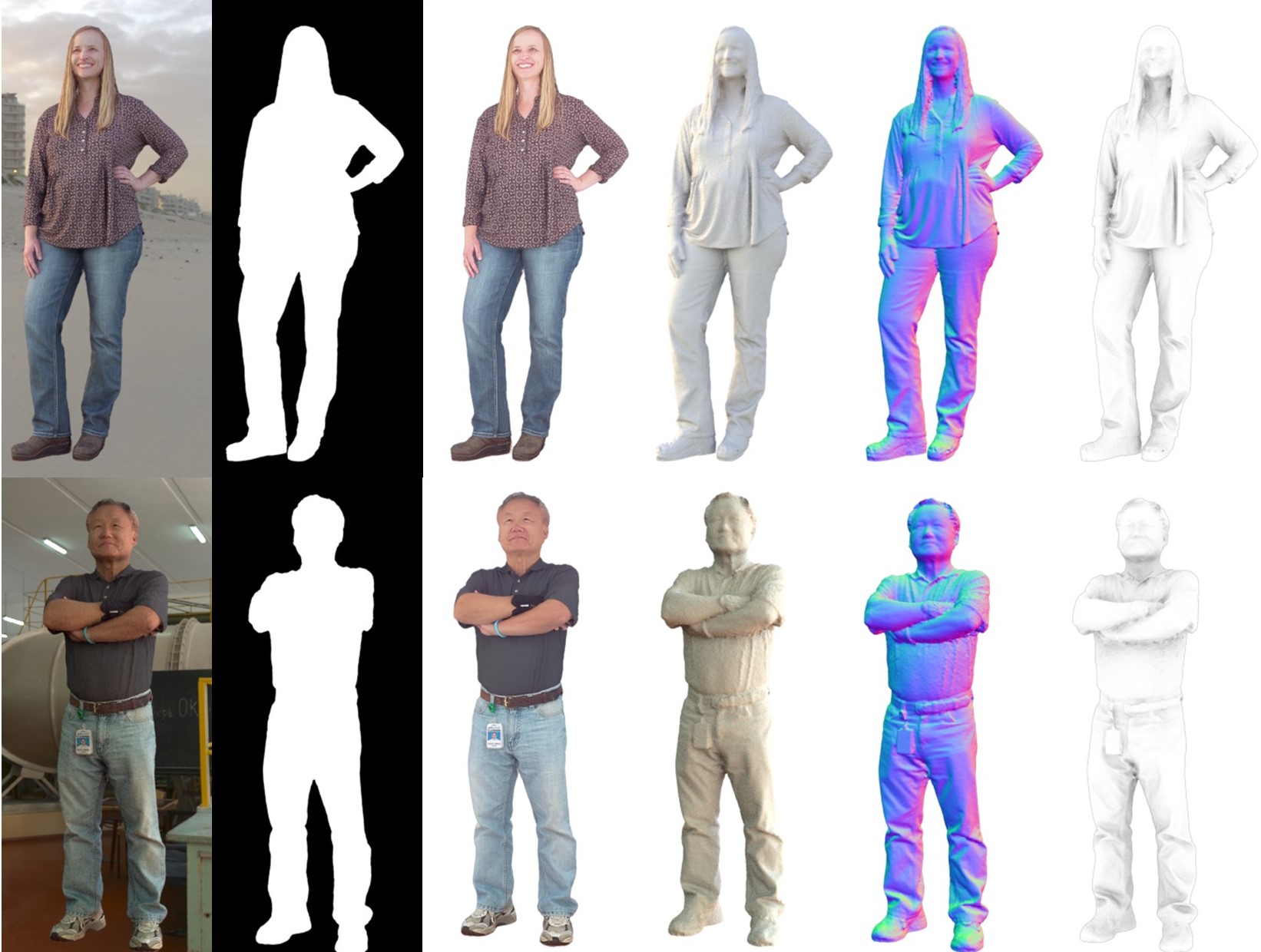} 
\caption{Examples from our human image dataset. For each human model, from left to right are the rendered image, mask, albedo map, shading map, normal map and ambient occlusion map.}
\label{fig:datset}
% \vspace{-0.5cm}
\end{figure}

We use Cycle rendering engine in Blender \cite{blender.org} and a Principled BSDF shader to render dataset.
For each model collected from Twindom \cite{twindom}, we set the pitch angle in [0,10,20,30], yaw angle in [-32,-24,-16,-8,0,8,16,24,32], and model scale in [0.8,1,1.1] and render the model under two different random lighting conditions for every setting. As shown in Fig.~\ref{fig:datset}, each model can produce 216 sets of images where each set consists of the rendered image, albedo map, normal map, ambient occlusion map and corresponding mask. The human models are collected with well-behaved lighting conditions and the color of the texture is treated as ground truth albedo. To obtain the ambient occlusion map, we add extra output node in Cycles engine \cite{blender.org} and enable ``AO pass'' during rendering. In total, we generate 150k sets of images for training and 20k sets of images for testing. The resolution of the rendered images is $512 \times 512$. All training data are augmented using random blur, noise and color enhancement. The Twindom dataset we used is restricted to make public and we will release pretrained model based on the THUman 2.0 dataset \cite{tao2021function4d} for research purposes.

\subsection{Training and Testing Details}
We train geometry networks, de-lighting nerwork and shading refinement networks separately in PyTorch. For all models that require training, we use the Adam optimizer and set learning rate to 1e-5. The learning rate for all discriminators is set to 1e-6. The size of inputs for geometry networks, de-lighting network and full-body shading refinement network is $512 \times 512$. For the face refinement network, all inputs are resized to $128 \times 128$.

All the training processes are conducted on a 4-RTX2080Tis-Server, which consists of: 1) 3D reconstruction using PIFuHD \cite{saito2020pifuhd}, 3DMM fitting \cite{guo2020towards} and ray-tracing take 2 days in total and 2) Training of Geometry Module, Albedo Module and Shading refinement module (body/face) takes 2 days, 4 days and 2 days respectively. The testing (on a single RTX2080Ti PC) efficiency is 17s per image which includes: 120 ms for estimating albedo, 14s for 3D reconstruction, 3s for ray-tracing and 50 ms for shading refinement.

For every 3D human model, there are two sets of rendered images under different lighting conditions for every pose. When testing, we use one set of images as input and predict the relit image under the lighting condition of another set. Therefore, we can compare the difference between the inferred relit result and the ground truth.

\subsection{Ray-traced 3D Model Smoothing}

\begin{figure}
\centering
\includegraphics[width=0.60\textwidth]{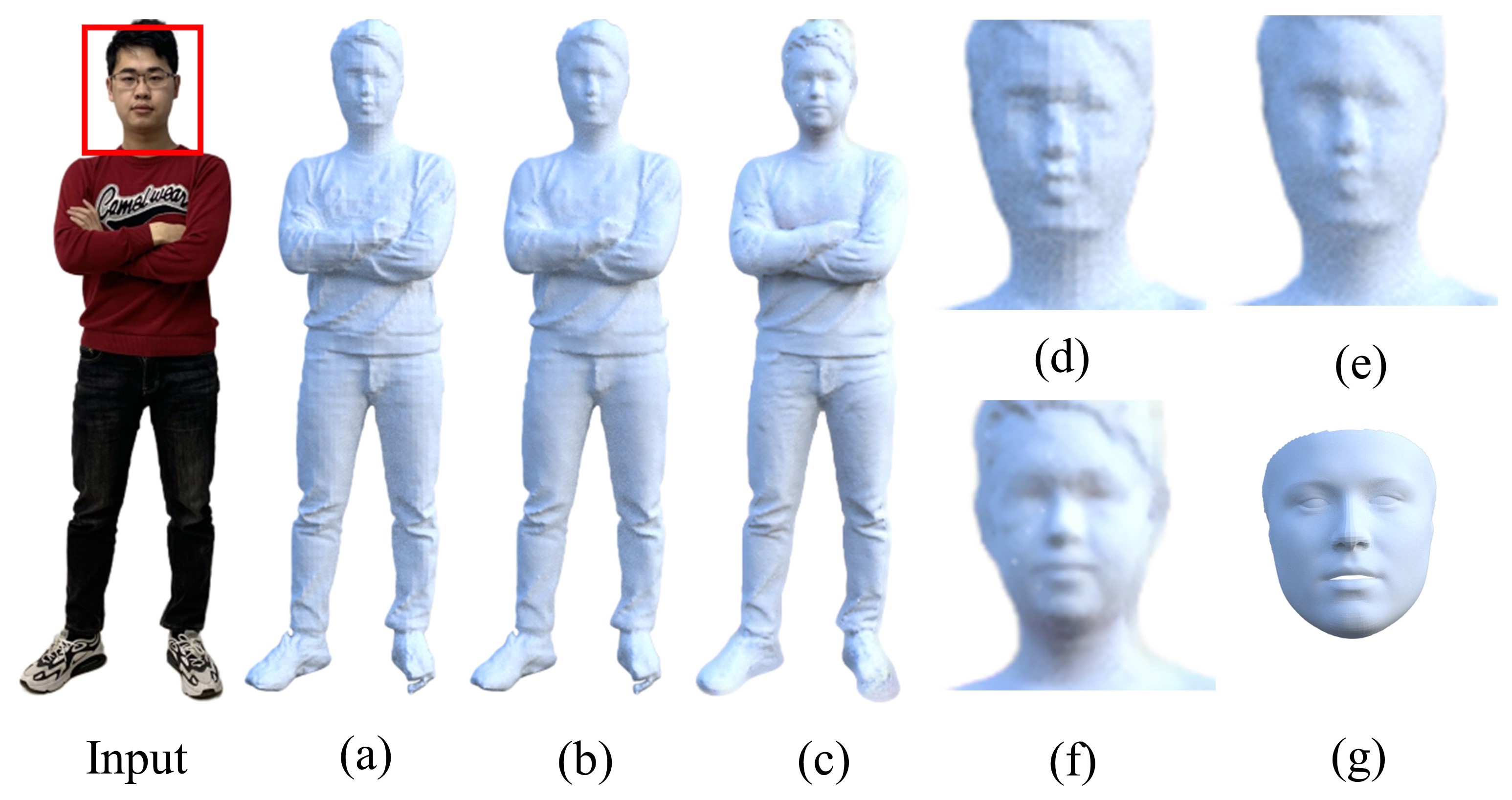} 
\caption{Ray tracing results under the target illumination condition for the 3D model estimated from the input image. (a) Ray-traced shading map without smoothing (b) Ray-traced shading map with smoothing (c) Refined shading map (d) Cropped face from the ray-traced shading map without smoothing (e) Cropped face from the ray-traced shading map with smoothing (f) Cropped face from the refined shading map (g) ray-traced 3DMM face shading map}
\label{fig:raytracing_abalation}
% \vspace{-0.4cm}
\end{figure}
The Fig.~\ref{fig:raytracing_abalation} shows the checkered artifacts mentioned in Section 5.1. The resolution of the estimated 3D model by PIFuHD \cite{saito2020pifuhd} is low, and direct ray tracing may produce grid-like artifacts. We use Laplacian smoothing to smooth the surface of the estimated full-body model and improve the quality of the ray-traced shading map. The ray-traced 3DMM face shading map is shown in Fig.~\ref{fig:raytracing_abalation} (g). Compared with cropped faces from the ray-traced full-body shading maps, it owns clear facial geometry details and shadows.

\subsection{Details of Comparisons with SOTA Methods \cite{Kanamori,tajimaPG21,Manuel}}
All input images are resized to $512 \times 512$. 
%However, they use different lighting representations. 
Since all three methods use spherical harmonic lighting representation (but with different orders), we extract 25 SH coefficients for every HDR environmental lighting map, and use the first 9 SH coefficients for testing RH \cite{Kanamori} and RHW \cite{tajimaPG21} and all 25 SH coefficients for testing SFHR \cite{Manuel}.

For the comparison with RH \cite{Kanamori}, we re-render our dataset using spherical harmonics lighting and pre-computed radiance transfer(PRT) to ensure that all the 3D models and lighting are the same as in our dataset.
Experimentally we find that the model of RH \cite{Kanamori} retrained on our larger dataset demonstrates superior performance, as shown in Tab.\ref{tab:Quantitative_evaluations}. 
% For a fair comparison, the qualitative results of \cite{Kanamori} come from the retrained model rather than the official pre-trained model.
For a fair comparison, the qualitative results of RH \cite{Kanamori} come from both the retrained model and the official pretrained model.
% For the comparison with SFHR\cite{Manuel}, the author generously provided the testing code and the pretrained model, but the corresponding dataset has not yet been released. Since our dataset does not support rendering highlight maps and residual maps, 
We directly test the released models of RHW \cite{tajimaPG21} and SFHR \cite{Manuel} on our testing dataset. Since these three methods require the brightness of target lighting to lie within [0.7,0.9], the relit results will fail under harsh illuminations and large areas of white pixels appear in the relit shading map. Therefore, we scale the pixel values of their relit shading to [0,1] by dividing the relit shading by its maximum value instead of clipping it to [0,1].

\section{Comparison with TotalRelighting \cite{pandey2021total}}
\begin{figure}[t]
\centering
\includegraphics[width=0.98\textwidth]{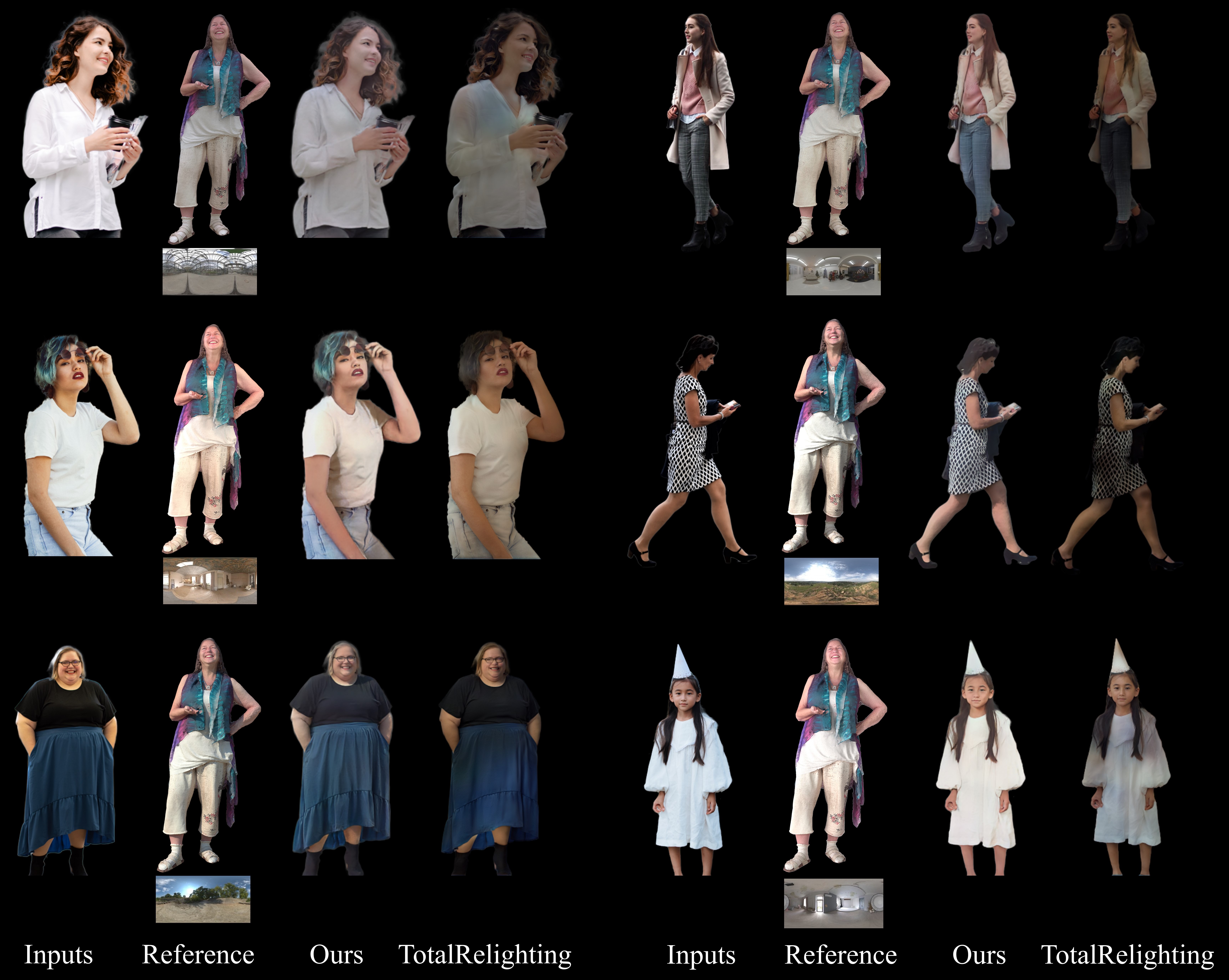} 
\caption{Relit results on real-world images. ''Reference'' are the rendered images of a virtual 3D human model under the target lighting conditions and are used to indicate the position of shadows. The target HDR environment map is placed under the reference image.}
\label{fig:tr}
% \vspace{-0.5cm}
\end{figure}

We show results in comparison with TotalRelighting \cite{pandey2021total}, which is trained on real OLAT dataset. As shown in Fig.~\ref{fig:tr}, the TotalRelighting may produce patchy shadows and inaccurate albedo inference for clothing. By contrast, our method is able to generate photo-realistic relit results with high-frequency self-shadows (including hard cast shadows).

\clearpage

\end{document}